\documentclass[]{article}

\usepackage{lineno,hyperref}
\usepackage{amsmath}
\usepackage{amsfonts}
\usepackage{amssymb}
\usepackage{calligra}
\usepackage{calrsfs}
\usepackage{todonotes}
\usepackage{makecell}
\usepackage{colortbl}
\usepackage[linesnumbered,lined,boxed,commentsnumbered]{algorithm2e}
\usepackage{euscript}
\usepackage{cancel}

\newcolumntype{R}[2]{%
	>{\adjustbox{angle=#1,lap=\width-(#2)}\bgroup}%
	l%
	<{\egroup}%
}

\tikzset{
	diagonal fill/.style 2 args={fill=#2, path picture={%
			\fill[#1] (path picture bounding box.south west) -|
			(path picture bounding box.north east) -- cycle;}},
	reversed diagonal fill/.style 2 args={fill=#2, path picture={
			\fill[#1] (path picture bounding box.north west) |- 
			(path picture bounding box.south east) -- cycle;}}
}

\newcommand{\bicl}[1]
{
	\cellcolor[gray]{0}\pitt\textcolor{white}{#1}
}

\newcommand{\pitt}
{
	\hspace*{-1.3ex}\rule{0pt}{2.3ex}
}

\newcommand{\ol}[1]{\overline{#1}}

\newtheorem{definition}{Definition}

\newcommand{\mytheoremonename}{Implicants and constant patterns}
\newcommand{\mytheoremtwoname}{Prime implicants and inclu\-sion--ma\-ximal constant patterns}
\newcommand{\mytheoremthreename}{Implicants and $\delta$--shifting patterns}
\newcommand{\mytheoremfourname}{Prime implicants and inclu\-sion--ma\-ximal $\delta$--shif\-ting patterns}
\newcommand{\mytheoremfivename}{Implicants and exhaustive search of all $\delta$--sensible shifting patterns}
\newcommand{\mytheoremsixname}{Prime implicants and exhaustive search of all inclusion--maximal $\delta$--sensible shif\-ting patterns}
\newcommand{\mytheoremsixnamep}{Prime implicants and ex\-haus\-tive  search of all inclusion--maximal $\delta$--sensible shif\-ting patterns}
\newcommand{\mytheoremsevenname}{Implicants and pruned search of all $\delta$--sensible shifting patterns}
\newcommand{\mytheoremeightname}{Prime implicants and pruned search of all inclu\-sion--maximal $\delta$--sensible shifting patterns}

\newcommand{\calm}{{\EuScript M}}
\newcommand{\calc}{{\EuScript C}}
\newcommand{\calr}{{\EuScript R}}

\newcommand{\cala}{{\EuScript A}}

\newcommand{\calb}{{\EuScript B}}
\newcommand{\cali}{{\EuScript I}}
\newcommand{\calj}{{\EuScript J}}

\newcommand{\implicant}{\EuScript P}
\newcommand{\ffunction}{\mathrm f}
\newcommand{\calh}{{\EuScript H}}

\title{Boolean Reasoning--Based Biclustering\\ for Shifting Pattern Extraction}
\author{Marcin Michalak$^1$, Jes\'{u}s S. Aguilar--Ruiz$^{2,3}$\\ \ \\ \normalsize{$^1$Department of Computer Networks and Systems,} \\ \normalsize{Silesian University of Technology}\\ \normalsize{ul. Akademicka 16, 44-100 Gliwice, Poland}\\ \normalsize{Marcin.Michalak@polsl.pl}\\ \ \\\normalsize{$^2$School of Engineering, Pablo de Olavide University,} \\ \normalsize{ES--41013 Seville, Spain}\\\normalsize{aguilar@upo.es}\\ \ \\\normalsize{$^3$Computer Science, Universidad Nacional de San Agust\'{i}n,} \\ \normalsize{Arequipa, Per\'{u}}}

\begin{document}

\maketitle

\begin{abstract}
Biclustering is a powerful approach to search for patterns in data, as it can be driven by a function that measures the quality of diverse types of patterns of interest. However, due to its computational complexity, the exploration of the search space is usually guided by an algorithmic strategy, sometimes introducing random factors that simplify the computational cost (e.g. greedy search or evolutionary computation). 

Shifting patterns are specially interesting as they account constant fluctuations in data, i.e. they capture situations in which all the values in the pattern move up or down for one dimension maintaining the range amplitude for all the dimensions. This behaviour is very common in nature, e.g. in the analysis of gene expression data, where a subset of genes might go up or down for a subset of patients or experimental conditions, identifying functionally coherent categories.  

Boolean reasoning was recently revealed as an appropriate methodology to address the search for constant biclusters. In this work, this direction is extended to search for more general biclusters that include shifting patterns. The mathematical foundations are described in order to associate Boolean concepts with shifting patterns, and the methodology is presented to show that the induction of shifting patterns by means of Boolean reasoning is due to the ability of finding all inclusion--maximal $\delta$--shifting patterns.

Experiments with a real dataset show the potential of our approach at finding biclusters with $\delta$--shifting patterns, which have been evaluated with the mean squared residue (MSR), providing an excellent performance at finding results very close to zero. 
\end{abstract}

\section{Introduction}
Biclustering is a two--dimensional clustering technique considered first by Morgan and Sonquist \cite{Morgan63}, and subsequently by Hartigan \cite{Hartigan72} and by Mirkin \cite{Mirkin96}, although it was popularized by Cheng and Church \cite{Cheng00} in the context of gene expression data analysis. It has an interesting property that is not fulfilled by clustering techniques: a value from the matrix (dataset) could belong to zero, one or more biclusters. In other words, the join of all the biclusters might not be the original dataset, and the intersection of two biclusters might not be empty. In many contexts, biclustering results are more appropriate, as they focus on more specific patterns (subsets of rows and columns, simultaneously) than those provided by clustering (row or column segmentation, independently). Apart from the fact that the computational complexity is higher than that of clustering (exponential), becoming a NP--hard problem \cite{Tanay2002}, biclustering has captured the attention of the scientific community because it is able to search for more specific patterns in data, discarding naturally what is not relevant for the goals \cite{PONTES2015163}. 

In general, the extraction of biclusters from data is structured in two processes: the search for biclusters in the high--dimensional search space (exploration), and the selection of good biclusters by means of a quality function (exploitation). The goal is to find a sub--matrix (or a set of) that shows a behavioral pattern for rows and columns simultaneously. Patterns can adopt different structures, depending on the definition of behavior associated to the values in the bicluster \cite{Madeira2004}. However, there are two types of patterns that show special interest in nature: shifting and scaling patterns \cite{Aguilar05}. Most of approaches for shifting pattern induction are based on a  measure named Mean Square Residue (MSR) \cite{Cheng00} and its variants \cite{Ahmed,Bryan2,Bryan,Reiss}, or on graph theory \cite{gracob,BiClusO,QUBIC}.

During the last two decades, there have been developed different heuristics that provide sets of biclusters:  based on graphs \cite{Tanay2002,DENITTO2017114,DENITTO2017186}, uncovering structures by eigenvectors \cite{gerstein03}, based on evolutionary computation \cite{Aguilar-Ruiz05c, Mitra06,  Divina_Aguilar06, Pontes09, PontesGA13}, ensemble methods \cite{HANCZAR20123938}, or scatter search \cite{Nepomuceno11, Nepomuceno15a}, which are evaluated with some measure of quality \cite{Banerjee07, GUPTA20102692,Divina12, Nepomuceno15b, Pontes2015,Nepomuceno2018,FLORES2013367}. However, they mainly focused on the type of bicluster (as defined in \cite{Madeira2004}) rather than on the type of pattern \cite{Aguilar05}.

The process of searching for biclusters in data can be addressed with different paradigms. In the formal concept analysis realm \cite{wille}, the extraction of the concept lattice is equivalent to finding inclusion--maximal biclusters in binary data. Similarly, the bicluster of ones in binary data  \cite{Rodriguez11}, that refers to market basket analysis \cite{Serin} (a special case of affinity analysis \cite{Aguinis}), may be easily interpreted as frequent item sets.

This work presents a new approach for pattern extraction based on Boolean reasoning \cite{BR}, and extends both the general biclustering approach started in \cite{michalak_slezak_biklastry} for binary and discrete data, and the later extended for continuous data \cite{michalak_slezak_cbiklastry}. The rationale behind the approach is based on the possibility of representing data differences as Boolean formulas, whose implicants (and prime implicants) encode the biclusters (such correspondence between biclusters and implicants is mathematically proven). In fact, shifting pattern induction is possible since Boolean reasoning provides the skills for finding all inclusion--maximal $\delta$--shifting patterns.

The paper is organized as follows: Sec. \ref{sec:definitions} provides the most important concepts used throughout the paper; Sec. \ref{sec:boolean} introduces the Boolean reasoning paradigm in the context of biclustering; the inference of patterns is presented in Sec. \ref{sec:pattern_induction}, including constant and shifting patterns, and then the $\delta$--shifting pattern; empirical results of the approach are shown in Sec. \ref{sec:results}; finally, conclusions and future work are summarized. In addition, mathematical proofs of all pattern induction theorems are shown in the Appendix.

\section{Definitions}\label{sec:definitions}
This section provides definitions of the most important concepts used throughout the document, together with some brief explanations.


Let $\ffunction$ be a Boolean function over a set $A$ of $v$ variables, that is, $A=\{a_1, \ldots$, $a_v\}$, and let $A'\subseteq A$ be a subset of variables.


\begin{definition}
	\label{def:def}
\textbf{(Conjunctive Normal Form (CNF))} The function $\ffunction$ is in Conjunctive Normal Form if it consists of a conjunction of clauses formed by disjunctions of simple or negated literals. The following example function is in CNF  (the overline symbol means the negation):
$$
\ffunction(\{a, b, c\}) = (a \lor b)\land (b \lor c) \land (\ol{a} \lor b)
$$
\end{definition}

\begin{definition}
\textbf{(Disjunctive Normal Form (DNF))} The function $\ffunction$ is in Disjunctive Normal Form if it consists of a disjunction of simple or negated literals. The following example function is in DNF:
$$
\ffunction(\{a, b, c\}) = (a \land b) \lor (a \land c) \lor (b \land \ol{c})
$$
\end{definition}



\begin{definition}\label{def:implicant}
\textbf{(Implicant)}
The expression $\implicant_\ffunction(A')$ is an implicant of $\ffunction(A)$ iff
$$
\implicant_\ffunction(A')=1\Rightarrow \ffunction(A)=1
$$
\noindent{where $A' \subseteq A$}.
\end{definition}

\begin{definition}\label{def:prime_implicant}
\textbf{(Prime Implicant)} An implicant $\implicant_\ffunction(A')$ (where $A' \subseteq A$) of a  function $\ffunction$ is a \emph{prime implicant} of $\ffunction$ iff there is no a proper subset $A'' \subset A'$ such that $$\implicant_\ffunction(A'')=1 \Rightarrow \ffunction(A) = 1$$
\end{definition}

When the DNF is the disjunction of (only) prime implicants then is also called Blake Canonical Form (BNF) \cite{BCF}.

The previous definitions will support the Boolean logic used in the methodology to represent potential biclusters. Boolean reasoning will help finding biclusters that meet a predefined quality criteria. Next, some definitions related to biclustering will be provided, so as the properties of biclusters that identify types of patterns in data.

\begin{definition}
\textbf{(Matrix or Dataset)}
A \textit{matrix} is defined as a tuple $\calm = (\calr, \calc)$, where
$\calr,\calc$ are two finite sets referred to as the \textit{set of rows} and the \textit{set of columns}, respectively.

A matrix $\calm = (\calr , \calc)$, is therefore defined by
$\calr=\{r_1$, $\dots$, $r_n\}, n \geq 1$, $\calc=\{c_1,\dots,c_m\}, m \geq 1$, where each position $(i,j) \in \calr \times \calc$ contains a real value:
\[
\left(v_{i,j}\right) = 
                 \left(
                 \begin{array}{ccc}
                  v_{1,1} &  \cdots & v_{1,m}\\
                  \vdots & \ddots & \vdots\\
                  v_{n,1} &  \cdots & v_{n,m}
                  \end{array}
                  \right)
\]
\end{definition}

Commonly, the indexes of $\calr$ and $\calc$ are natural numbers. However, in this paper those natural indexes will be interpreted as nominal labels, so as the inclusion relation of row/column indexes is implicit.



\begin{definition}\label{def:bicluster_ii}
\textbf{(Bicluster)} A \textit{bicluster} is a 2--tuple $\calb$ $=$ $(\cali$, $\calj)$=$\left(w_{i,j}\right)$, where $\cali \subseteq
\calr$, $\calj \subseteq \calc$, and $\forall i\in \cali$ and $\forall j\in \calj$ it satisfies $w_{i,j} = v_{i,j}$.
\end{definition}

Hereafter, for simplicity, the bicluster $\calb$ will be represented as a matrix of values $\left(w_{i,j}\right)$, or enumerating the subset of rows and columns as $(\cali,\calj)$, and generically, with the expression $\cali\calj$. As $\cali$ and $\calj$ are subsets of $\calr$ and $\calc$, respectively, empty biclusters will be identified when any of those subsets is empty.

A bicluster consists of a subset of rows and a subset of columns such that the rows exhibit similar behavior across the columns, or vice versa. The problem of discovering a set of biclusters from a matrix is conditioned by some specific measure of homogeneity (quality), which assigns better values to biclusters that identify valid patterns in data. 

\begin{definition}
\textbf{(Perfect Bicluster)} 
A perfect bicluster is a submatrix $(\cali , \calj)$ that satisfies one of the following:
\begin{itemize}
\item All the values are equal, i.e., a constant $\pi$.
\item All the values within the bicluster can be obtained using one of the following
expressions:
\begin{equation}\label{eq:cons_cond_add}
w_{i,j}= \pi+\alpha_i
\end{equation}
\begin{equation}\label{eq:cons_cond_mul}
w_{i,j}= \pi \times \alpha_i
\end{equation}
where $\pi$ is a typical value within the bicluster and $\alpha_i$ is the adjustment
for row $i \in \cali$. This adjustment can be obtained either in an additive
(Eq. \ref{eq:cons_cond_add}) or multiplicative (Eq. \ref{eq:cons_cond_mul}) way. 
\item All the values within the bicluster can be obtained using one of the following
expressions:
\begin{equation}\label{eq:cons_gene_add}
w_{i,j}= \pi+\beta_j
\end{equation}
\begin{equation}\label{eq:cons_gene_mul}
w_{i,j}= \pi \times \beta_j
\end{equation}
where $\pi$ is a typical value within the bicluster and $\beta_j$
is the adjustment for column $j \in \calj$. This adjustment can be obtained
either in an additive (Eq. \ref{eq:cons_gene_add}) or multiplicative (Eq. \ref{eq:cons_gene_mul}) way.
\item All the values within the bicluster are coherent in both rows and columns, so each value $w_{ij}$ can be obtained using one of the following expressions:
\begin{equation}\label{eq:coh_gene_add}
w_{i,j}= \pi+\alpha_i+\beta_j
\end{equation}
\begin{equation}\label{eq:coh_gene_mul}
w_{i,j}= \pi \times \alpha_i \times \beta_j
\end{equation}
\end{itemize}
\end{definition}

\begin{definition}\label{def:incl_max}
\textbf{(Inclusion--maximal Bicluster)} Let $\calb$ $=$ $(\cali$, $\calj)$ be a bicluster of the matrix $\calm$, whose elements satisfy some well defined criterion $\calh$.  The bicluster $\calb$ is inclusion--maximal iff there is neither $i \in \calr\setminus\cali$ nor $j \in \calc\setminus \calj$ such that any extended biclusters $\calb_i = (\cali \cup \{i\}, \calj)$ or $\calb_j =(\cali, \calj \cup \{j\})$ still satisfies the criterion $\calh$. In other words, subsets $\cali$ and $\calj$ are maximal subsets (in the sense of inclusion) of sets $\calr$  and $\calc$, respectively.
\end{definition}

According to Def. \ref{def:incl_max} there exist at most two inclusion--maximal empty biclusters in the given matrix: the first consists of all the rows and no columns, while the second consists of no rows and all the columns. It refers to two situations, respectively:
\begin{itemize}
    \item there is no such a column whose all the elements fulfill the criterion $\calh$,
    \item there is no such a row whose all the elements fulfill the criterion $\calh$.
\end{itemize}

\begin{definition}\label{def:shifting_pattern}
\textbf{(Shifting Pattern)} A bicluster $(\cali , \calj)$ shows a \emph{shifting pattern} when it follows the expression:
\begin{equation}\label{eq:shifting}
w_{i,j}= \pi_j+\beta_i \hspace{1cm} \forall i\in \cali,\forall j\in \calj
\end{equation}
This definition introduces the non--constant values of $\pi$, i.e., $\pi_{j}$ could vary across columns, although maintaining the same value for all the rows within a column. 
\end{definition}

\begin{figure}[ht!]
  \centering
  \includegraphics[width=0.85\linewidth, trim = 80 20 70 0, clip]{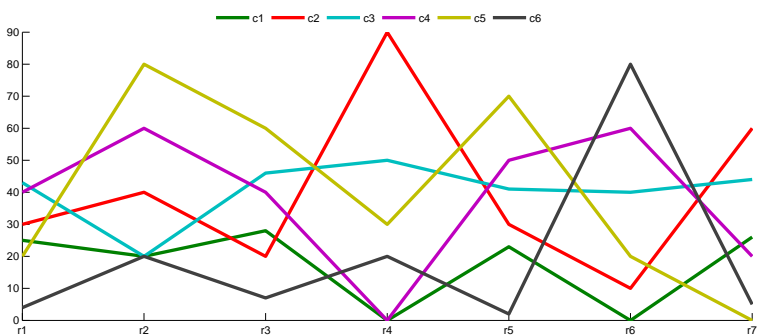}
  \caption{$\calm_0$ matrix presented as the set of column series.}
  \label{dataset}
\end{figure}

\begin{figure}[ht!]
  \centering
    \includegraphics[width=0.85\linewidth, trim = 80 20 70 0, clip]{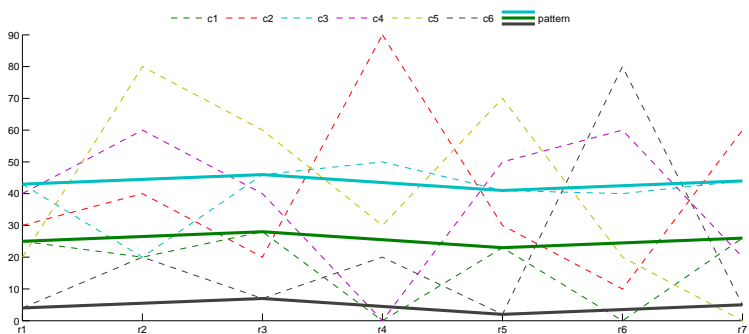}
  \caption{Shifting pattern presented over the original $\calm_0$ data.}
  \label{shifting}
\end{figure}

The matrix $\calm_0$, with seven rows and six columns (Table \ref{tab:matrix_0}), is depicted in  Fig. \ref{dataset}. Each column $c$ of the data is presented as the line going through rows indicated in the $X$ axis. Fig. \ref{shifting} reveals a hidden shifting pattern $(\{r_1, r_3$, $r_5, r_7\}$, $\{c_1, c_3, c_6\})$ over the original data (drawn with dotted lines).

\begin{table}[!ht]

    \centering \scriptsize
    \begin{tabular}{c|c|c|c|c|c|c|}
	&	$c_1$	&	$c_2$	&	$c_3$	&	$c_4$	&	$c_5$	&	$c_6$	\\ \hline
$r_1$	&	25	&	30	&	43	&	40	&	20	&	4	\\
$r_2$	&	20	&	40	&	20	&	60	&	80	&	20	\\
$r_3$	&	28	&	20	&	46	&	40	&	60	&	7	\\
$r_4$	&	0	&	90	&	50	&	0	&	30	&	20	\\
$r_5$	&	23	&	30	&	41	&	50	&	70	&	2	\\
$r_6$	&	0	&	10	&	40	&	60	&	20	&	80	\\
$r_7$	&	26	&	60	&	44	&	20	&	0	&	5	\\ \hline
    \end{tabular}
     \caption{A sample matrix $\calm_0$.}
    \label{tab:matrix_0}
\end{table}

In real domains it is very unlikely to find shifting patterns, as data always has inherent noise --- although usually very small. Only an infinitesimal variation of a matrix value would cause the row and column where it is placed to not be both selected for a bicluster.

\begin{definition}{Definition}
\textbf{(Row/Column corresponding Boolean Variable)}
Let $\calm$ be a matrix of $n$ rows ($\calr = \{r_1,\ldots, r_n\}$) and $m$ columns ($\calc = \{c_1, \ldots, c_m\}$). Let $i \in \calr$ be a row of $\calm$, then $i'$ is its corresponding row Boolean variable. Similarly, let $j \in \calc$ be a column in $\calm$, then $j'$ is its corresponding column Boolean variable. 
\end{definition}

For the sake of simplicity, apostrophes ($'$) will be removed from row/column Boolean variables and the meaning will depend on the context (row/column or Boolean variable associated to row/column, respectively).

\begin{definition}\label{def:bicluster_andimplicant}
\textbf{(Bicluster and Implicant correspondence)}
Let $\calm$ be a given matrix of $n$ rows ($\{r_1,\ldots, r_n\}$) and $m$ columns ($\{c_1, \ldots, c_m\}$). 
Let $\calb$ $=$ $(\cali$, $\calj)$ be a bicluster of $\calm$. The implicant  $\implicant(\calb)$ is called the $\cali\calj$ corresponding bicluster iff it contains only the Boolean variables that correspond to rows $r\in \calr$ such that $r\notin \cali$ and columns $c\in \calc$ such that $c\notin \calj$. 
\end{definition}

The corresponding implicant will be denoted as $$\implicant(\calb(\cali, \calj))=(\cali', \calj')$$ or just shorter $\cali'\calj'$. Different notations refer to the context defined in Defs. \ref{def:implicant} and \ref{def:prime_implicant}. There, no distinction among types of Boolean variables was provided. Here, the implicant (or the prime implicant) is built from two types of Boolean variables, whose interpretation differs significantly: each Boolean variable from $\cali'$ is a row corresponding variable while each Boolean variable from $\calj'$ is a column corresponding variable. Such a distinction will be very helpful when further definitions, theorems and proofs are provided.


For example, the bicluster $\cali\calj = (\{r_1, r_3, r_6\}$, $\{c_1, c_2, c_3\})$ from matrix $\calm_0$ (Table \ref{tab:matrix_0}) would have the corresponding implicant $\cali'\calj'$ $ = r_2 \land r_4 \land r_5 \land r_7  \land c_4 \land c_5 \land c_6$.


\section{Boolean Reasoning in Biclustering}\label{sec:boolean}
The first applications of the Boolean reasoning paradigm in the domain of biclustering were published by Michalak and \'Sl\c{e}zak \cite{michalak_slezak_biklastry}, in where mathematical foundations were provided for discrete value matrices \cite{UitertMW08,Rodriguez11}. The intuitive generalization of such an approach for continuous data was later presented in \cite{michalak_slezak_cbiklastry}. However, both previous approaches do not deal with shifting patterns, as those addressed the search for global patterns, which fluctuate within a range (a global bandwidth). This work, instead, introduces a Boolean reasoning--based approach to search for $\delta$--shifting patterns, which are much more interesting due to their usefulness in many domains (for instance, gene expression data analysis \cite{Aguilar05}).

In order to illustrate how the Boolean reasoning can help finding biclusters, a simple example will be shown. The maximal absolute difference between any two cells in matrix $\calm_1$ (Table \ref{michalak_slezak:tab:cM}) is equal to 4. The greatest absolute difference smaller than 4 is 3. The only pair of cells whose difference exceeds 3 is $v_{1,1}=1$ and $v_{2,3}=5$. The CNF clause that codes these cells coordinates with row and column corresponding Boolean variables is $(r_1 \lor c_1 \lor r_2 \lor c_3)$. This clause is also already in DNF and consists of four prime implicants.

\begin{table}[ht!]
	\centering \scriptsize

	\begin{tabular}{c|c|c|c|}	
		& $c_1$ & $c_2$ & $c_3$ \\ \hline
		$r_1$ & 1 & 2 & 3 \\ 
		$r_2$ & 2 & 3 & 5 \\ 
	\end{tabular}
		\caption{A sample matrix $\calm_1$.}
	\label{michalak_slezak:tab:cM}
\end{table}

It was proved in \cite{michalak_slezak_cbiklastry} that biclusters associated to prime implicants are defined by the variables that are not present in the prime implicants. From matrix $\calm_1$ and the function $\ffunction = (r_1 \lor c_1 \lor r_2 \lor c_3)$ (pairs exceeding the value 3), several biclusters can be provided, as shown in Table \ref{michalak_slezak:tab:M_res3}. 

\begin{table}[ht!]
	\centering \scriptsize

	\begin{tabular}{cc}
		\begin{tabular}{c|c|c|c|}
			
			& $c_1$ & $c_2$ & $c_3$ \\ \hline
			$r_1$ & 1 & 2 & 3 \\ 
			$r_2$ & \bicl{2} & \bicl{3} & \bicl{5} \\ 
		\end{tabular}
		& 
		\begin{tabular}{c|c|c|c|}
			
			& $c_1$ & $c_2$ & $c_3$ \\ \hline
			$r_1$ & 1 & \bicl{2} & \bicl{3} \\ 
			$r_2$ & 2 & \bicl{3} & \bicl{5} \\ 
		\end{tabular}
		\\
		$r_1 : (\{r_2\},\{c_1, c_2, c_3\})$	
		&
		$c_1 : (\{r_1,r_2\},\{c_2, c_3\})$			
		\\
		\begin{tabular}{c|c|c|c|}
			
			& $c_1$ & $c_2$ & $c_3$ \\ \hline
			$r_1$ & \bicl{1} & \bicl{2} & \bicl{3} \\ 
			$r_2$ & 2 & 3 & 5 \\ 
		\end{tabular}
		&
		\begin{tabular}{c|c|c|c|}
			
			& $c_1$ & $c_2$ & $c_3$ \\ \hline
			$r_1$ & \bicl{1} & \bicl{2} & 3 \\ 
			$r_2$ & \bicl{2} & \bicl{3} & 5 \\ 
		\end{tabular} \\

		$r_2 : (\{r_1\},\{c_1, c_2, c_3\})$	
		
		&	
		$c_3 : (\{r_1, r_2\},\{c_1, c_2\})$

	\end{tabular}
		\caption{Four biclusters identified by the prime implicants, and the corresponding bicluster representation $\cali\calj$ for matrix $\calm_1$.}\label{michalak_slezak:tab:M_res3}
\end{table}

For example, the first implicant is $r_1$, that corresponds to the bicluster formed by all the columns and only the row $r_2$. Also, there is {neither column nor row that can be added to the bicluster without violating the defined property on maximal absolute difference. The example reveals that it is possible to express global properties of biclusters in terms of Boolean reasoning. However, as it was introduced in Sec. \ref{sec:definitions}, it becomes interesting to analyse only in--row absolute differences instead of global ones, as it implicitly considers the order of rows, what has important consequences for the analysis of data in several domains (e.g. venereal tumors \cite{Chokeshaiusaha2019}, time--lagged expression data \cite{Madeira2014}).

The previous approach was successfully developed for finding constant biclusters in discrete and binary data \cite{michalak_slezak_biklastry} so as for finding biclusters of similar values in continuous data \cite{michalak:bib:michalak_icmmi,michalak_slezak_cbiklastry}. This work extends the research to address the search for biclusters in continuous data that include more sophisticated patterns (not only constant real values), and it requires new definitions, proofs and the methodology to validate the approach in the context of real--world problems.

\section{Pattern Induction with Boolean Reasoning}\label{sec:pattern_induction}

The definitions to support the procedure of extracting biclusters in real--valued domains by means of Boolean reasoning will be presented next.

\subsection{Constant Patterns}\label{sec:constant_patterns}

\begin{definition}\label{def:exact_pattern}
\textbf{(Constant pattern)} The $\cali\calj$ bicluster is a constant pattern of matrix $\calm$ iff $\forall i \in \cali$ and $\forall j, k \in \calj$ it satisfies $v_{i,j} = v_{i,k}$.
\end{definition}

Table \ref{tab:exact_pattern} presents the matrix $\calm_2$ of continuous values, containing the constant pattern $(\{r_1,r_2,r_4,r_5\}, \{c_1, c_2\})$ which have the same value for a subset of columns. Moreover, there are neither rows nor columns that can be added without violating the condition of inclusion--maximality (see Def. \ref{def:incl_max}).

\begin{table}[ht!]
	\centering \scriptsize
	\begin{tabular}{c|c|c|c|}
		  & $c_1$ & $c_2$ & $c_3$ \\ \hline
		$r_1$ & \bicl{1} & \bicl{1} & 2 \\ 
		$r_2$ & \bicl{2} & \bicl{2} & 4 \\ 
		$r_3$ & 3 & 1 & 5 \\ 
		$r_4$ & \bicl{4} & \bicl{4} & 5 \\ 
		$r_5$ & \bicl{5} & \bicl{5} & 1 \\ 
		$r_6$ & 2 & 1 & 5 \\ \hline
	\end{tabular}
	\caption{A sample matrix $\calm_2$ with a constant pattern.}\label{tab:exact_pattern}
\end{table}

Boolean reasoning can help finding all the patterns by defining the function that encodes all the pairs of cells (at each row independently) with different value.

\begin{definition}\label{def:bool_differences}
\textbf{(Boolean function that encodes all in--row absolute differences)} Let $\calm$ be a matrix of rows $\calr$ and columns $\calc$. The Boolean function $\ffunction$ that encodes all in--row pairs with different values is defined as follows:
$$
\ffunction(\calm) = \bigwedge(i \lor j \lor k)
$$
\noindent where
$$
i \in \calr, \quad j,k \in \calc
$$
\noindent such that
$$
v_{i,j} \neq v_{i,k}
$$
\end{definition}

For the matrix $\calm_2$ in Table \ref{tab:exact_pattern} the Boolean function $\ffunction$ is:
$$
\ffunction(\calm_2) = (r_1 \lor c_1 \lor c_3)\land(r_1 \lor c_2 \lor c_3)\land(r_2 \lor c_1 \lor c_3)\land(r_2 \lor c_2 \lor c_3)\land
$$
$$
\quad\  \land(r_3 \lor c_1 \lor c_2)\land(r_3 \lor c_1 \lor c_3)\land(r_3 \lor c_2 \lor c_3)\land(r_4 \lor c_1 \lor c_3)\land
$$
$$
\quad\ \land(r_4 \lor c_2 \lor c_3)\land(r_5 \lor c_1 \lor c_3)\land(r_5 \lor c_2 \lor c_3)\land(r_6 \lor c_1 \lor c_2)\land
$$
$$
\land(r_6 \lor c_1 \lor c_3)\land(r_6 \lor c_2 \lor c_3)
$$
\noindent and simplifying, in DNF:
$$\ffunction(\calm_2) = (c_1\land c_2) \lor (r_1\land r_2\land r_3\land r_4\land r_5\land r_6) \lor 
$$
$$
\lor (r_3\land r_6 \land c_3) \lor (c_1\land c_3) \lor (c_2\land c_3)
$$
The final form of the $\ffunction(\calm_2)$ has additionally introduced brackets for the better presentation of its prime implicants. 

Prime implicants of $\ffunction(\calm_2)$ function encode the patterns in such a way that the pattern consists of both rows and columns whose corresponding Boolean variables are not present in the prime implicant.

The constant pattern shown in Table \ref{tab:exact_pattern} is easily identified by means of its prime implicant ($r_3\land r_6\land c_3$), which is associated to the bicluster $(\{r_1,r_2,r_4,r_5\},$  $\{c_1, c_2\})$, in accordance to the Def. \ref{def:bicluster_andimplicant}. The first implicant $(c_1\land c_2)$ and the last two $(c_1\land c_3)$ and $(c_2 \land c_3)$ refer to single--column patterns. It might not be so intuitive that each of column represents a pattern that only contains one column with all rows -although formally correct-. Finally, the second implicant ($r_1\land r_2\land r_3\land r_4\land r_5\land r_6$) refers to the empty pattern. Such empty biclusters are consistent with the Def. \ref{def:incl_max} and their interpretation is very important. The bicluster that consists of all columns and no rows (like the bicluster that corresponds to the prime implicant above) means that there is no such constant pattern in data that covers at least one whole row. However, such empty bicluster still fulfills the assumed criterion and it is inclusion--maximal (no column nor row might be added).

The theorems that establish the relationship between implicants and constant patterns, and between prime implicants and inclusion--maximal constant patterns, respectively, are enunciated next.

\newtheorem{mythrm}{Theorem}\label{th:exact:correctness}
\begin{mythrm}
\textbf{(\mytheoremonename)} $\cali'\calj'$ is an implicant of $\ffunction_{ep}(\calm)$ iff $\cali\calj$ is a constant pattern in $\calm$.
\end{mythrm}

\begin{mythrm}\label{th:exact:maximality}
\textbf{(\mytheoremtwoname)}
	$\cali'\calj'$ is a prime implicant of $\ffunction_{ep}(\calm)$ iff $\cali\calj$ is an inclusion--maximal constant pattern in $\calm$.
\end{mythrm}

\subsection{$\delta$--Shifting Patterns}\label{sec:delta_shifting}

The concept of shifting pattern was introduced in Def. \ref{def:shifting_pattern}, and it can be conveniently weakened in order to consider that the maximal absolute in--row difference between pairs of cells will not exceed a given threshold $\delta$.

\begin{definition}\label{def:d_shifting_pattern}
\textbf{($\delta$--Shifting Pattern)} 
A bicluster $\cali\calj$ shows a \emph{$\delta$--shifting pattern} when: \begin{equation}\label{eq:delta_shifting}
\forall i\in \cali,\ \forall j,k \in \calj \hspace{1cm} \left|v_{i,j}-v_{i,k}\right|\leq \delta
\end{equation}
\end{definition}

It becomes intuitive (from the relationship between Defs. \ref{def:exact_pattern} and \ref{def:bool_differences}) that proper Boolean functions should be based on the negation of the condition in Eq. \ref{eq:delta_shifting} --- the encoding of all in--row pairs that violate this condition. 

\begin{definition}\label{def:threshold_shifting_pattern}
\textbf{(Boolean function that encodes all in--row absolute differences greater than $\delta$)}
Let $\calm$ be a matrix of rows $\calr$ and columns $\calc$. The Boolean function $\ffunction_{\delta}$ that encodes all in--row pairs whose difference is not greater than $\delta$ is defined as follows:
$$
\ffunction_{\delta}(\calm) = \bigwedge(i \lor j \lor k)
$$
\noindent where
$$
i \in \calr, \quad j,k \in \calc
$$
\noindent such that
$$
|v_{i,j} - v_{i,k}| > \delta
$$
\end{definition}

\begin{table}[h]
	\centering \scriptsize
	\begin{tabular}{c|c|c|c|}
		  & $c_1$ & $c_2$ & $c_3$ \\ \hline
		$r_1$ & 1 & 3 & 2 \\ 
		$r_2$ & 1 & 3 & 4 \\ 
		$r_3$ & 2 & 1 & 5 \\ \hline
	\end{tabular}
	\caption{A sample matrix $\calm_3$.}\label{tab:matrix_m2}
\end{table} 

Table \ref{tab:matrix_m2} shows the matrix $\calm_3$ with three rows and three columns. To build the $\ffunction_{\delta=2}$ formula (to find inclusion--maximal $\delta$--shifting patterns) , all the pairs of cells whose absolute difference exceeds the value 2 should be found (at each row separately). The first row does not contain such pairs so it does not generate any clause. There is one pair in the second row $v_{2,1}$ and $v_{2,3}$, whose absolute difference is 3. Such pair is encoded with a disjunction of three Boolean variables that correspond to the row ($2$) and to the columns ($1$ and $3$) so the clause has the following form: $r_2 \lor c_1 \lor c_3$. All clauses are logically multiplied so the final CNF expression looks as follows:
$$
\ffunction_{\delta = 2}(\calm_3) = (r_2 \lor c_1 \lor c_3)\land (r_3 \lor c_1 \lor c_3) \land (r_3 \lor c_2 \lor c_3)
$$

Transforming the CNF function into DNF would discover the presence of interesting prime implicants encoding $\delta$--shifting patterns:
$$
\ffunction_{\delta = 2}(\calm_3)  =   (r_2 \lor c_1 \lor c_3)\land (r_3 \lor c_1 \lor c_3) \land (r_3 \lor c_2 \lor c_3) =
$$
$$
  =   (r_2 \land r_3 \lor c_1 \lor c_3) \land (r_3 \lor c_2 \lor c_3)  =
$$
$$
  =  (r_2 \land r_3) \lor (r_3 \land c_1) \lor (c_1 \land c_2) \lor \cancel{c_1 \land c_3 } \lor \cancel{r_3 \land c_3} \lor \cancel{c_2 \land c_3} \lor c_3
$$

Finally, four prime implicants were found (Table \ref{fig:m2:results}) associated to biclusters that are inclusion--maximal $\delta$--shifting patterns, and whose row differences do not exceed 2. 

\begin{table}[ht!]
	\centering \scriptsize
	\begin{tabular}{ccc}
			\begin{tabular}{c|c|c|c|}
			& $c_1$ & $c_2$ & $c_3$ \\ \hline
			$r_1$ & \bicl{1} & \bicl{3} & \bicl{2} \\ 
			$r_2$ & 1 & 3 & 4 \\ 
			$r_3$ & 2 & 1 & 5 \\ 
		\end{tabular}
	& & 
		\begin{tabular}{c|c|c|c|}
		& $c_1$ & $c_2$ & $c_3$ \\ \hline
		$r_1$ & 1 & \bicl{3} & \bicl{2} \\ 
		$r_2$ & 1 & \bicl{3} & \bicl{4} \\ 
		$r_3$ & 2 & 1 & 5 \\ 
	\end{tabular}
\\
$r_2 \land r_3 : (\{r_1\}, \{c_1, c_2, c_3\})$
& &
$r_3\land c_1 : (\{r_1, r_2\}, \{c_2, c_3\})$
 \\
 & & \\
	\begin{tabular}{c|c|c|c|}
	& $c_1$ & $c_2$ & $c_3$ \\ \hline
	$r_1$ & 1 & 3 & \bicl{2} \\ 
	$r_2$ & 1 & 3 & \bicl{4} \\ 
	$r_3$ & 2 & 1 & \bicl{5} \\ 
\end{tabular}
& &
	\begin{tabular}{c|c|c|c|}
	& $c_1$ & $c_2$ & $c_3$ \\ \hline
	$r_1$ & \bicl{1} & \bicl{3} & 2 \\ 
	$r_2$ & \bicl{1} & \bicl{3} & 4 \\ 
	$r_3$ & \bicl{2} & \bicl{1} & 5 \\ 
	
\end{tabular}\\
$c_1\land c_2 : (\{r_1, r_2, r_3\}, \{c_3\})$
& &
$c_3 : (\{r_1, r_2, r_3\}, \{c_1, c_2\})$ 
	\end{tabular}

	\caption{Visualization of prime implicants of function $\ffunction_{\delta = 2}(\calm_3)$.}\label{fig:m2:results}
\end{table}

In short, it is possible to express the $\delta$--shifting pattern induction in terms of Boolean reasoning--based functions, which is demonstrated in Ths. \ref{th1} and \ref{th2}.


\begin{mythrm}\label{th1}
\textbf{(\mytheoremthreename)}
	$\cali'\calj'$ is an implicant of $\ffunction_\delta(\calm)$ iff $\cali\calj$ is a $\delta$--shifting pattern in $\calm$.
\end{mythrm}

\begin{mythrm}\label{th2}
\textbf{(\mytheoremfourname)}
	$\cali'\calj'$ is a prime implicant of $\ffunction_\delta(\calm)$ iff $\cali\calj$ is an inclusion--maximal  $\delta$--shifting pattern in $\calm$.
\end{mythrm}

The presented approach limits the search of patterns, since it depends on the choice of the $\delta$ value. The next subsection will provide strategies that avoid such constraint.

\subsubsection{Exhaustive Search}\label{lbl:exhaustive}

The previous section introduced the searching for $\delta$--shifting patterns in data in terms of Boolean reasoning. However, the search requires the in--row variability level ($\delta$) assumption within biclusters. The choice of the value of $\delta$ is therefore critical, and it might provide high number of small patterns or low number of large patterns depending of its fluctuation. The solution to this issue is to consider as many values of $\delta$ as possible.

Apparently, the $\delta$ value could be chosen from the continuous range (up to the maximal absolute in--row difference). Fortunately, ``sensible'' $\delta$ values can be identified thus helping reduce the search space complexity. 

Let the ordered set of  all the $q$ possible in--row differences in some matrix  $\{ \alpha_1, \ldots, \alpha_q\}$ be considered, such that $i<j \Rightarrow \alpha_i < \alpha_j$. The pattern induction results for any $\delta=\alpha_k$, where $\alpha_k \in \{ \alpha_1, \ldots, \alpha_q\}$, should be different than those for $\delta = \alpha_{k+1}$. However, any given $\gamma \in [\alpha_k, \alpha_{k+1})$ would lead to the same results with $\delta = \gamma$ as the obtained for $\delta = \alpha_k$. For that reason, the values from $\alpha_1$ to $\alpha_q$ are called ``sensible'' with respect to the $\delta=\alpha_k$ value. All the differences from the input data can be encoded in one formula, what allows to provide all inclusion--maximal $\delta$--shifting patterns for each $\delta$.

\begin{definition}\label{def:ordered_set}
\textbf{(Ordered set of sensible differences)}
Let $\calm$ be a matrix. The ordered set of increasing absolute in--row differences is defined as follows:
$$
\Delta(\calm) = \{\alpha_1, \ldots, \alpha_{q}\}
$$
\noindent where $\forall i, j$  such that $1\leq i< j \leq q $ it is implied that $\alpha_i < \alpha_j$.
\end{definition}

\begin{definition}\label{def:sensible_difference}
\textbf{(Sensible difference corresponding Boolean variable)}
Let the matrix $\calm$ and its ordered set of sensible differences  $\Delta(\calm)$ be given. For each element $\alpha$ of the set $\Delta(\calm)$, the Boolean variable $\alpha'$ is called sensible difference corresponding Boolean variable.
\end{definition}

As previously mentioned, the apostrophe distinguishes between sensible absolute differences ($\alpha$: numerical variable) and its corresponding Boolean variable ($\alpha'$), although throughout the text the apostrophe will be omitted in Boolean formulas for simplicity, and therefore the meaning of $\alpha$ will be context dependent. It is also possible to introduce two additional definitions for row/column indices and $\alpha$ level mapping functions, thus increasing the paper formality, although for the sake of simplicity this step has been omitted.

\begin{definition}\label{def:delta_shifting_pattern}
\textbf{(Boolean function that encodes all sensible absolute in--row differences)}
Let $\calm$ be a matrix of rows $\calr$ and columns $\calc$. The Boolean function $\ffunction_\alpha$ that encodes all in--row pairs of different values given an absolute difference $\alpha$ is defined as: 
$$
\ffunction_\alpha(\calm) = \bigwedge(i \lor j \lor k \lor \alpha)
$$
\noindent where
$$
i \in \calr, \ j,k \in \calc, \ \alpha \in \Delta(\calm)
$$
\noindent  such that
$$
|v_{i,j} - v_{i,k}| \geq \alpha
$$
\end{definition}

\begin{table}[ht]
    \centering \scriptsize
    \begin{tabular}{c|c|c|} 
         & $c_1$ & $c_2$ \\ \hline
        $r_1$ & 1.0 & 1.0\\ \hline 
        $r_2$ & 2.0 & 4.3\\ \hline 
        $r_3$ & 2.0 & 3.2\\ \hline 
        $r_4$ & 2.0 & 5.1\\ \hline 
    \end{tabular}
    \caption{The sample matrix $\calm_4$.}
    \label{tab:m3}
\end{table}

Table \ref{tab:m3} illustrates an example of using the set of absolute differences $\Delta(\calm_4)$ (Def. \ref{def:ordered_set}) and how to encode all sensible absolute in--row differences by a Boolean function (Def. \ref{def:delta_shifting_pattern}). There is no difference between cells in the first row, while in the second, third and fourth the row differences are equal to 2.3, 1.2 and 3.1, respectively. The ordered set $\Delta(\calm)$ is as follows: 
$$
\Delta(\calm_4) = \{1.2, 2.3, 3.1\} = \{\alpha_1, \alpha_2, \alpha_3\}
$$

Finally, according to Def. \ref{def:delta_shifting_pattern}:
$$
\ffunction_{\alpha}(\calm_4) = (r_2 \lor c_1 \lor c_2\lor \alpha_1)\land(r_2 \lor c_1 \lor c_2\lor \alpha_2)\land(r_3\lor c_1\lor c_2\lor\alpha_1)\land
$$
$$
\land(r_4 \lor c_1 \lor c_2 \lor\alpha_1)\land(r_4 \lor c_1 \lor c_2 \lor\alpha_2)\land(r_4 \lor c_1 \lor c_2 \lor\alpha_3)
$$
\noindent and transformed into DNF:
$$
\ffunction_{\alpha}(\calm_4) = c_1 \lor c_2 \lor (r_2\land r_3\land r_4) \lor (r_2\land r_4\land\alpha_1) \lor 
$$
$$
\lor (r_4\land\alpha_1\land\alpha_2) \lor  (\alpha_1\land\alpha_2\land\alpha_3)
$$

Intuitively, each prime implicant should correspond to some inclusion--maximal shifting pattern in the matrix. The interpretation of prime implicants without $\alpha$ variables is quite obvious: the implicants $c_1$ and $c_2$ are single column constant patterns, while the implicant ($r_2\land r_3\land r_4$) refers to the whole first row, which is also a constant pattern. 

\begin{table}[h]
    \centering \scriptsize
    \begin{tabular}{ccc}
    \begin{tabular}{c|c|c|} 
         & $c_1$ & $c_2$ \\ \hline
        $r_1$ & \bicl{1.0} & \bicl{1.0}\\ \hline 
        $r_2$ & 2.0 & 4.3\\ \hline 
        $r_3$ & \bicl{2.0} & \bicl{3.2}\\ \hline 
        $r_4$ & 2.0 & 5.1\\ \hline 
    \end{tabular}
    & &
        \begin{tabular}{c|c|c|} 
         & $c_1$ & $c_2$ \\ \hline
        $r_1$ & \bicl{1.0} & \bicl{1.0}\\ \hline 
        $r_2$ & \bicl{2.0} & \bicl{4.3}\\ \hline 
        $r_3$ & \bicl{2.0} & \bicl{3.2}\\ \hline 
        $r_4$ & 2.0 & 5.1\\ \hline 
    \end{tabular}
    \end{tabular}
    \caption{The sample matrix $\calm_4$ and two patterns corresponding to prime implicants: $r_2\land r_4\land \alpha_1$ (on the left) and $r_4\land\alpha_1\land\alpha_2$ (on the right). The first prime implicant corresponding bicluster is $(\{r_1, r_3\}, \{c_1, c_2\})$ and the second corresponding bicluster is $(\{r_1, r_2, r_3\}, \{c_1, c_2\})$.}
    \label{tab:m3:alpha1}
\end{table}

Table \ref{tab:m3:alpha1} shows the implicant $(r_2\land r_4\land \alpha_1)$, whose in--row absolute difference does not exceed $\alpha_1$. Similarly, the implicant $(r_4\land\alpha_1\land\alpha_2)$ does not exceed $\alpha_2$ (the biggest $\alpha$ absolute difference corresponding Boolean variable). 

As the exhaustive search introduces new notions, the following definitions will help present the theorems related to exhaustive search of patterns.

\begin{definition}
\textbf{(Subset of consecutive sensible differences)}
Let the matrix $\calm$ and the ordered set of sensible differences $\Delta(\calm)$ be given, then the set
$$ 
\cala_p = \{\alpha_k \in \Delta(\calm) : \alpha_k \leq \alpha_p , \forall k \in \{1, \ldots, q\} \} 
$$
\noindent is called the subset of consecutive sensible differences.
\end{definition}

\begin{definition}
\textbf{(Set of $\cala_p$ elements corresponding Boolean variables)}
Let the matrix $\calm$ and the ordered set of sensible differences $\Delta(\calm)$ be given, then the set of Boolean variables
$$
\cala'_p = \{\alpha' :  \alpha \in \Delta(\calm)-\cala_p\}
$$
\noindent is called a set of $\cala_p$ elements corresponding variables. 
\end{definition}

In other words $\cala'_p$ contains only those $\alpha'$ corresponding Boolean variables whose associated absolute difference in $\Delta(\calm)$ is not present in $\cala_p$ (according to Def. \ref{def:sensible_difference}).

\begin{definition}
\textbf{(Pattern and implicant correspondence)}
Let $\calm$ be a matrix of $n$ rows ($\{r_1, \ldots, r_n\}$), $m$ columns $(\{c_1, \ldots, c_m\})$ and $q = |\Delta(\calm)|$ sensible differences . Let $\calb = (\cali, \calj)$ be a $\alpha_p$--shifting pattern in $\calm$. The 3--tuple $\cali'\calj'\cala'_p$ will be called an implicant corresponding to $\calb$ iff: a) $\cali'$ is the set of $\cali$ elements corresponding Boolean variables; b) $\calj'$ is the set of $\calj$ elements corresponding Boolean variables; c) $\cala'_p$ is the set of $\cala_p$ elements corresponding Boolean variables.
\end{definition}



\begin{mythrm}\label{th:exhaustive:implicant}
\textbf{(\mytheoremfivename)}
	$\cali'\calj'\cala'_p$ is an implicant of $\ffunction_\alpha(\calm)$ iff $\cali\calj$ is an $\alpha_p$--shifting pattern in $\calm$.
\end{mythrm}

\begin{mythrm}\label{th:exhaustive:prime_implicant}
\textbf{(\mytheoremsixname)}
	$\cali'\calj'\cala'_p$ is a prime implicant of $\ffunction_\alpha(\calm)$ iff $\cali\calj$ is an inclusion--maximal  $\alpha_p$--shifting pattern in $\calm$.
\end{mythrm}

The proofs of these theorems can be found in the Appendix.

\subsubsection{Pruned Strategy}

The Boolean reasoning--based procedure for $\delta$--shifting pattern induction, which in--row difference does not exceed a given level was proposed above. Next, the previous subsection provided the exhaustive strategy of searching for all possible (and sensible) $\delta$--shifting patterns in data. The first mentioned approach limits the searching space while the second one guarantees the complete exploration of the search space.

Both approaches can be joined in order to provide a set of patterns that fulfill two criteria: patterns do not exceed a given $\delta$ (and they are inclusion--maximal) and found patterns are all the possible to be found (there is no such pattern in data that was not found by the method). Moreover, this approach provides results that do not require postprocessing since $\delta$--shifting patterns with high $\delta$ were already removed.

The formula that encoded all sensible differences and led to the all sensible patterns from Table \ref{tab:m3} in CNF is:
$$
\ffunction_{\alpha}(\calm_3)  =  (r_2 \lor c_1 \lor c_2\lor \alpha_1)\land(r_2 \lor c_1 \lor c_2\lor \alpha_2)\land
$$
$$\land (r_3\lor c_1\lor c_2\lor\alpha_1) \land (r_4 \lor c_1 \lor c_2 \lor\alpha_1) \land$$
$$
\land (r_4 \lor c_1 \lor c_2 \lor\alpha_2)\land(r_4 \lor c_1 \lor c_2 \lor\alpha_3)
\
$$
\noindent while in DNF is as follows:
$$
\ffunction_{\alpha}(\calm_3) = c_1 \lor c_2 \lor (r_2\land r_3\land r_4) \lor (r_2\land r_4\land\alpha_1) \lor 
$$
$$
\lor (r_4\land\alpha_1\land\alpha_2) \lor  (\alpha_1\land\alpha_2\land\alpha_3)
$$
The last prime implicant encodes the pattern that consists of all rows and all columns with the assumption that the maximal in--row difference in the pattern does not exceed $\alpha_3$, which is also the global absolute in--row difference. 

Let us suppose that the maximal acceptable absolute in--row difference is $\alpha_1$.  The Boolean formula in CNF is: 
$$
(r_2 \lor c_1 \lor c_2) \land  (r_3\lor c_1\lor c_2\lor\alpha_1) \land (r_4 \lor c_1 \lor c_2)
$$
Cells from the first row are equal so there is no such a condition that their difference may violate and this pair has no influence on the formula. Cells from the second row always violates the $\alpha_1$ condition so their corresponding clause does not contain any difference corresponding variable. The same situation takes place for the cells from the fourth row, whose corresponding clause is the third one in the formula. Finally, the difference between the third row cells is equal to $\alpha_1$, which does not reach the maximal level of difference criterion, what implies that the level of difference is represented by the proper difference corresponding Boolean variable.
The DNF of this expression consists of four clauses:
$$
(r_2\land r_3\land r_4)\lor (r_2\land r_4\land\alpha_1) \lor c_1 \lor c_2
$$
There are no prime implicants that would encode any $\delta$--shifting pattern with $\delta > \alpha_1$. Only constant and $\alpha_1$--shifting patterns are found. The formal definition of that Boolean formula is presented below. 

\begin{definition}
\textbf{(Boolean function that encodes all sensible and upper limited in--row differences)}
Let $\calm$ be a matrix of rows $\calr$ and columns $\calc$. The Boolean function $\ffunction_{\alpha, \delta}$ that encodes all in--row pairs of different values, which absolute difference exceeds the level $\delta$, is defined as follows:
$$
\ffunction_{\alpha, \delta}(\calm) = \bigwedge \left\{
\begin{array}{ll}
    (i \lor j \lor k) & \textrm{iff}\ |v_{i,j} - v_{i,k}| > \delta \\
    (i \lor j \lor k \lor \alpha) & \textrm{iff}\ |v_{i,j} - v_{i,k}| \leq \alpha \leq \delta
\end{array}
\right.
$$
\noindent where
$$
i \in \calr, \ j, k \in \calc, \ \alpha \in \Delta(\calm)
$$
\end{definition}

The $\alpha$ index of the function refers to the exhaustive search, as it took place for $\ffunction_\alpha$ before, while $\delta$ index refers to the maximal acceptable in--row absolute difference for found patterns.

According to $\cala'_p$ set of literals defined in previous subsection, the following theorems provide the linkage between such pruned strategy of pattern induction and the proper Boolean function analysis.

\begin{mythrm}\label{th:exhaustive_limited:implicant}
\textbf{(\mytheoremsevenname)}
$\cali'\calj'\cala'_p$ is an implicant of $\ffunction_{\alpha,\delta}(\calm)$ iff $\cali\calj$ is a $\alpha_p$--shifting pattern in $\calm$ (or the constant pattern if $\cala_p' = \emptyset$).
\end{mythrm}

\begin{mythrm}\label{th:exhaustive_limited:prime_implicant}
\textbf{(\mytheoremeightname)}
	$\cali'\calj'\cala'_p$ is a prime implicant of $\ffunction_{\alpha,\delta}(\calm)$ iff $\cali\calj$ is an inclusion--maximal  $\alpha_p$--shifting pattern in $\calm$  (or the inclusion--maximal constant pattern if $\cala_p' = \emptyset$).
\end{mythrm}

Proofs of these two theorems can be found in the Appendix.

\section{Experimental Analysis}\label{sec:results}
In order to show the quality of results we have selected a well--know dataset related to central nervous system development, which consists of 9 conditions and 112 genes \cite{GEM}. Genes were clustered into related expression patterns to infer regulatory origins and interactions between families across the transition of the rat cervical spinal cord from a primary to a highly differentiate state, determined by embryonic days 11 through 21 (E11-E21), postnatal days 0-14 (P0-P14), and adult (A) at 90 days. This work was pioneer at suggesting that similarities in expression patterns might point to the existence of common regulatory structures, useful in defining roles for genes with unknown function.The choice of the dataset is justified by its features to show empirically the validity of the proposed methodology, focusing on the quality of the results from the analytical perspective, without elaborating on their biological interpretation.




A first analysis was done to visualize the distribution of all in--row absolute differences. Fig. \ref{fig:dist} shows the distribution (histogram) for all the 55,944 pairs, which provide 2,667 unique differences, ranging from 0 (minimum) to 27.69 (maximum), with a mean of 0.88. About 8\% of differences are not greater than 0.1, 15\% than 0.2, 22\% than 0.3, and almost 55\% does not exceed 1.0. This suggests that the range [0,0.4] for $\delta$ is very reasonable for further evaluation, as it is covering more than $25\%$ of pairs.


 \begin{figure}[hbt]
     \centering
     \includegraphics[width=\linewidth, trim = 90 20 80 20 , clip]{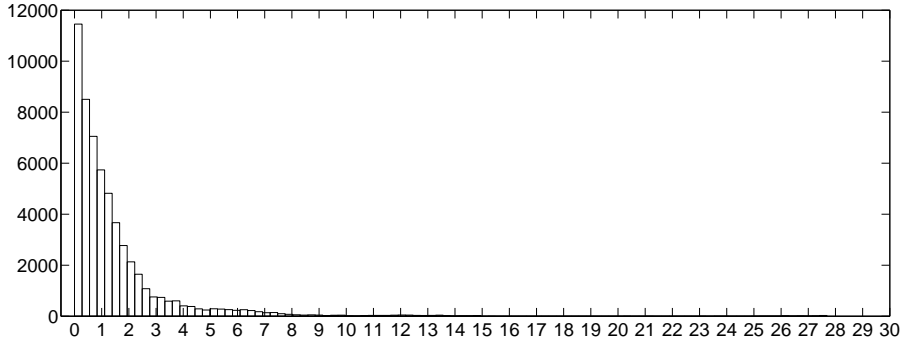}
     \caption{Histogram of all in--row absolute differences from data.}
     \label{fig:dist}
 \end{figure}

For $\delta=0$ a total of 257 constant patterns were discovered, including one empty pattern and 112 single--column ones (each one of them containing all the rows). These two special cases (empty and single--column) are always discovered --when they exist-- by the methodology, as Boolean reasoning also considers these patterns as biclusters (theoretically they are indeed). From this point, these patterns will be omitted for greater values of $\delta$, and we will focus on patterns with at least 2 rows and 2 columns, as they seem to be the smallest patterns that generalize interesting properties in both dimensions. Taking this into account, there are 27 constant patterns in data.

For $\delta=0.1$ there were discovered 1,027 patterns, and these can fluctuate within that difference, behaving as shifting patterns. For $\delta=0.2$ the number of patterns increases up to 2,487, for $\delta=0.3$ up to 4,027, and for $\delta=0.4$ up to 6,943. The quality of biclusters was measured to ensure that an increase in size does not necessarily lead to a decrease in quality. As balanced measure of size we will use the harmonic diameter ($d$), defined as $d(r,c)=\frac{2}{r^{-1}+c^{-1}}$, where $r$ and $c$ are the number of rows and  columns of the bicluster, respectively. This measure reflects partially the shape of the pattern. Patterns with similar (or comparable) areas may have quite different harmonic diameters. For instance, three patterns with area equal to 20 ($1\times 20, 2\times 10$ and $4\times 5$) will have their $d$ equal 1.9, 3.33 and 4.44, respectively. However, the third one has the highest generalization ability.

The mean squared residue (MSR) has been chosen as a measure of quality for biclusters because it is able to identify correctly constant and shifting patterns in data \cite{Cheng00,Aguilar05}. A low value (close to 0) means that the shifting pattern has no noise, and a high value means that either the shifting pattern has much noise or there is involved a more complex pattern in the bicluster (e.g. scaling pattern).

Table \ref{tab:results_summary} summarizes the results of biclusters containing shifting patterns obtained by varying the value of $\delta$. The maximum values of mean squared residue and harmonic diameter are also shown, and there is no significant loss of quality measured by the MSR while the value of $\delta$ increases.

 \begin{table}[tb]
     \centering \scriptsize
     \begin{tabular}{|c|c|c|c|}\hline
          $\delta$ & number of patterns & max(MSR) & $\max(d)$\\ \hline
          0.0 & \ \ \ \ 27 & 0.00000 & \ 4.20\\ 
          0.1 &  1,027 & 0.00226 & \ 5.45\\
          0.2 &  2,487 & 0.00951 & \ 7.00\\ 
          0.3 & 4,027 & 0.02102& \ 9.26\\ 
          0.4 & 6,943 & 0.02976 & 10.29\\ \hline
     \end{tabular}
     \caption{Summary of $\delta$--shifting patterns (not smaller than  $2\times 2$) provided for several values of $\delta$, including the highest value for the mean squared residue (worst case) and for the harmonic diameter.}
     \label{tab:results_summary}
 \end{table}


 
 
As only 27 meaningful patterns were found for $\delta=0$ (constant patterns), we have further limited the number of comparisons for the next values of $\delta$ with the goal of illustrating the good performance of the methodology regarding the size of biclusters (Fig. \ref{fig:best27_hd}). However, this increase in size has no negative effect in the quality of patterns, since the values of MSR oscillates very little when the harmonic diameter increases (Fig. \ref{fig:best27_msr}), which remains very close to zero. In Figs. \ref{fig:best27_hd} and \ref{fig:best27_msr}, the biclusters were decreasingly ordered by the harmonic diameter, so the X--axis represents the position of the bicluster in the ranking.
 
 \begin{figure}[t]
     \centering
     \includegraphics[width=\linewidth, trim = 70 0 60 0, clip]{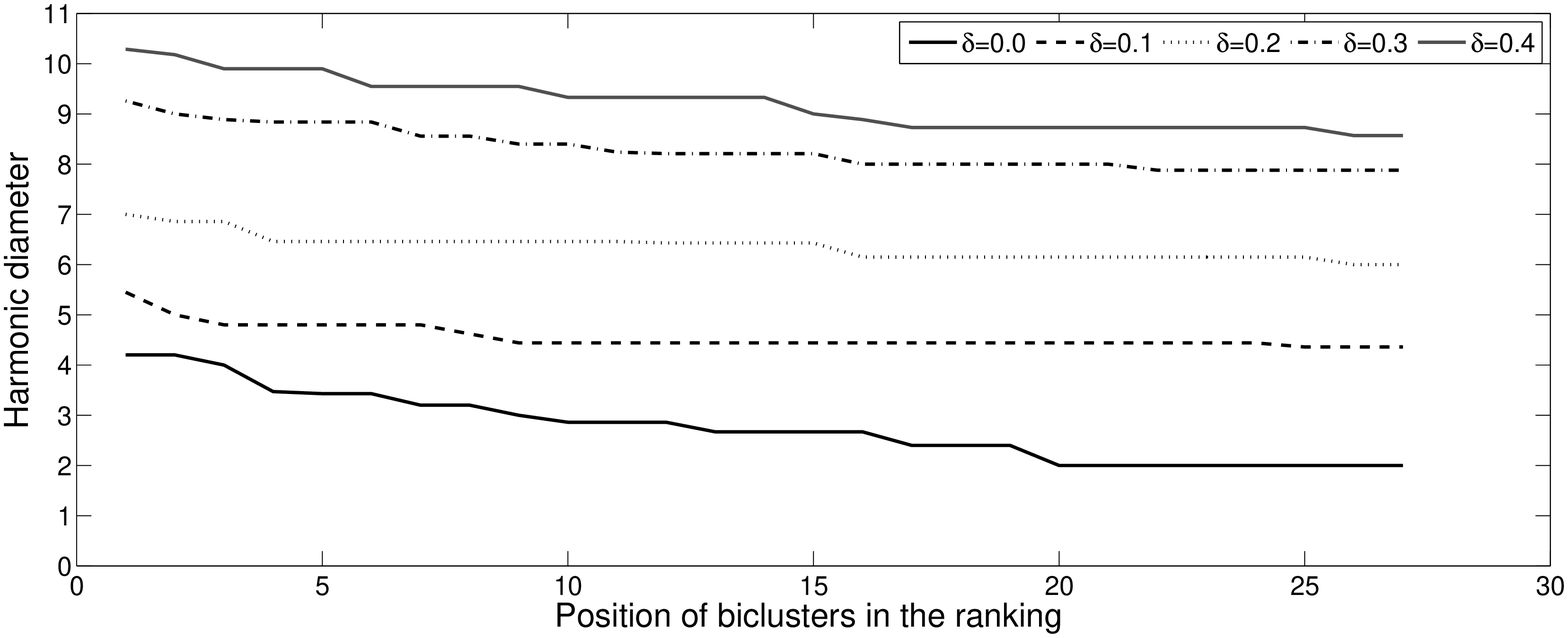}
     \caption{Harmonic diameter of the 27 patterns ranked with highest $d$ for every value of $\delta$.}
     \label{fig:best27_hd}
 \end{figure}
 
  \begin{figure}[t]
     \centering
     \includegraphics[width=\linewidth, trim = 50 0 50 0, clip]{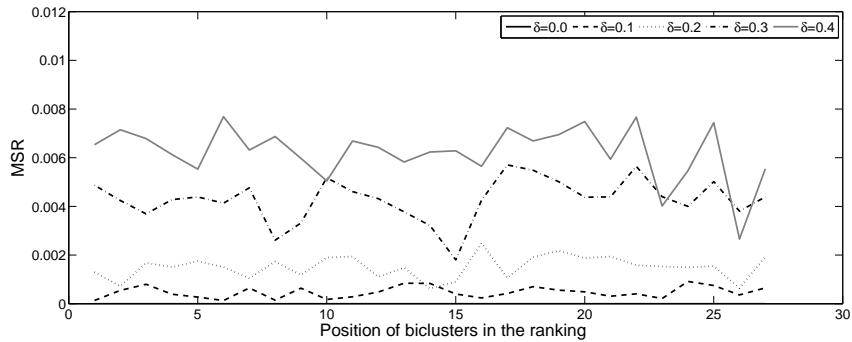}
     \caption{MSR of the 27 patterns ranked with highest $d$ for every value of $\delta$.}
     \label{fig:best27_msr}
 \end{figure}

There exists an stable relationship between MSR and $d$, since the variations in $d$ do not necessarily lead to the same behavior in MSR for $\delta=0.1$ (Fig. \ref{fig:hd_msr_01}). Thus, higher values of MSR do not correspond to higher values of $d$, what suggests that larger patterns still have consistent in--row values, and this also occurs for greater values of $\delta$.
 
   \begin{figure}[bt]
     \centering
     \includegraphics[width=\linewidth, trim = 60 0 70 0, clip]{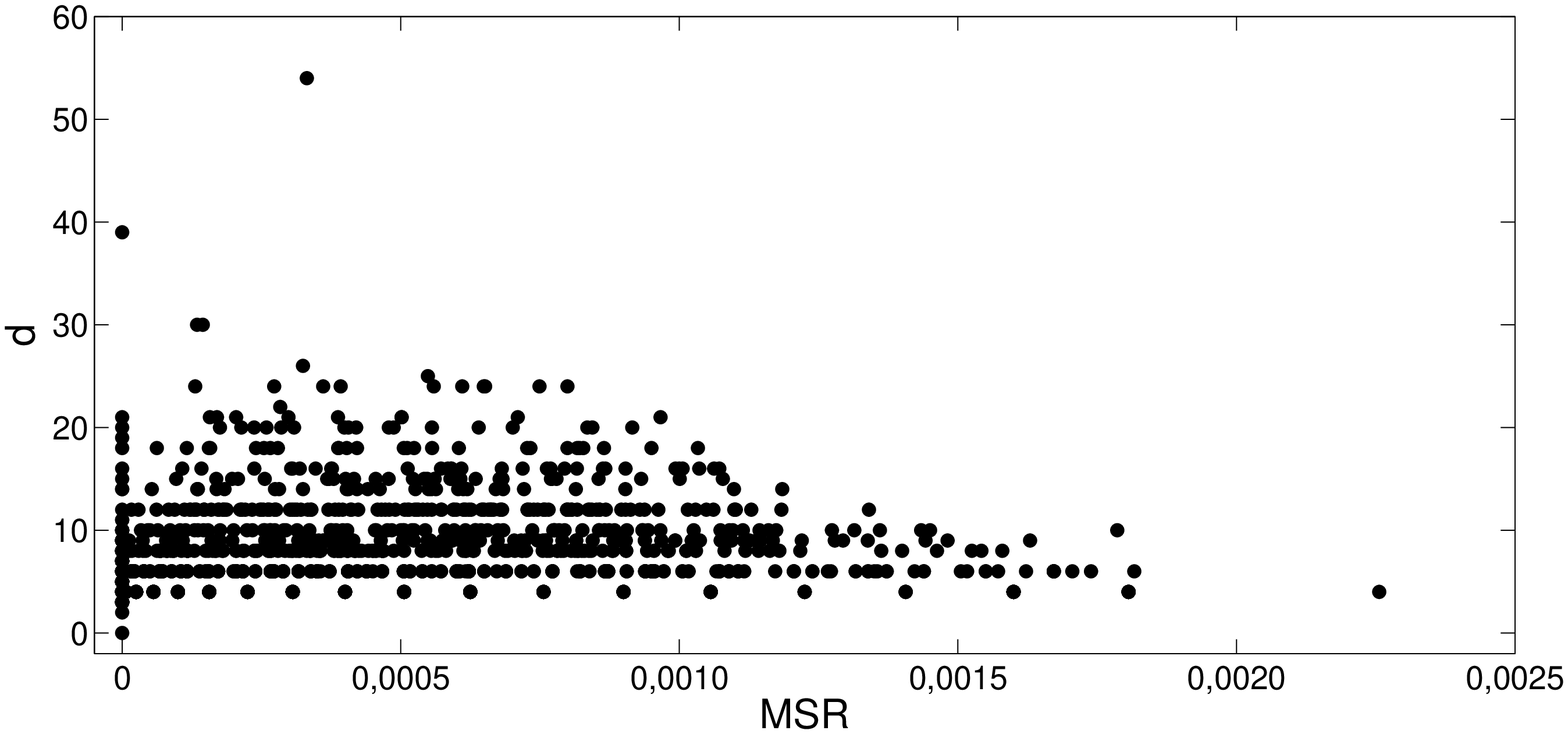}
     \caption{Relationship between the mean squared residue MSR and the harmonic diameter $d$ for 0.1--shifting patterns.}
     \label{fig:hd_msr_01}
 \end{figure}

 It is also worth to notice that $\delta =0.3$ allowed to find some patterns that went through all of the conditions and more than one gene: 12 patterns moved through 2 genes, 1 through 3, 2 through 4 and 7, and finally, 1 through 9 genes.  Greater increase of $\delta$ (0.4) provided 30 patterns: 16 going through 2 genes, 7 through 3 genes, 2 through 4 genes, and 1 through 5, 7, 8, 9 and 12 genes, respectively.
 



Examples of results for each value of $\delta$ will be shown next. Solid lines represent the behavior of each gene through the conditions. Additionally, the upper and lower bound of the pattern is marked with thicker dotted lines. The title of each subfigure indicates the maximal in--row difference of the pattern, which must be less than or equal to the $\delta$ value. A table is associated with each figure, in which details for each bicluster is presented.

 
  \begin{figure*}[!ht]
     \centering
     \includegraphics[width = \textwidth, trim = 130 30 110 10, clip]{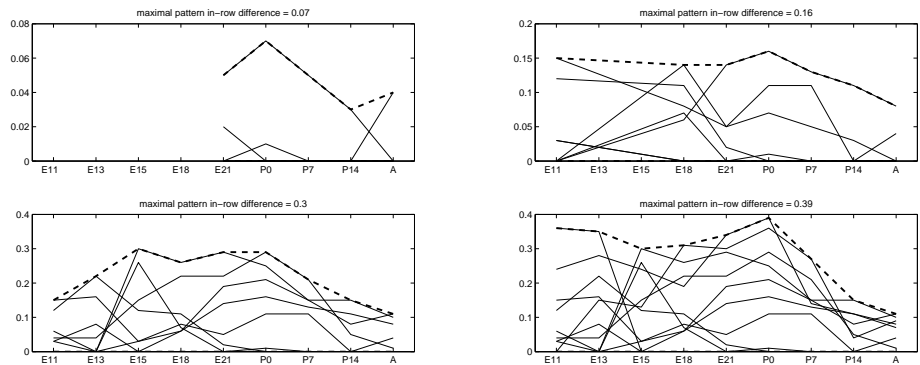}
     \caption{Best $\delta$--shifting patterns (in terms of harmonic diameter) found for several levels of $\delta$: $\delta = 0.1$ (upper left), $\delta = 0.2$ (upper right), $\delta=0.3$ (bottom left) and $\delta = 0.4$ (bottom right). Upper and lower bound of the pattern is marked with thick dotted lines, while gene values through conditions are represented with thin solid lines.}
     \label{fig:best_patterns}
 \end{figure*}
 
The first group (Fig. \ref{fig:best_patterns}) shows the best bicluster (in terms of MSR) for each value of $\delta \neq 0$. Complementary information on each bicluster is presented in Table \ref{tab:best_patterns}. Sizes of biclusters are slightly bigger as the value of $\delta$ increases, although with no significant impact on its quality, as the MSR remains very close to zero. In this case, all the values within biclusters are very low, so it is not possible to appreciate great oscillations in behavior.

The approach does not only find biclusters whose lower bounds of the conditions are close to 0. The second group (Fig. \ref{fig:nice_patterns}) shows other 4 biclusters with high fluctuations across the experimental conditions, which reveals important aspects related to gene regulation. The quality of biclusters (Table \ref{tab:nice_patterns}), measured by the MSR, still remains very close to 0.
 
It is important to highlight that the lowest MSR score indicates that the values fluctuate in unison, with constant or shifting patterns. The values of MSR calculated for the patterns presented in previous tables are extremely low, taking into account that for a bicluster with values randomly and uniformly generated in the range of $[l,u]$ the expected variance is $(l-u)^2/12$. The ranges for the conditions vary from [0,5.59] (E13) up to [0,27.69] (A), all of them starting from zero, so the expected MSR is much higher than those depicted in Table \ref{tab:results_summary}, and thereby in Tables \ref{tab:best_patterns} and \ref{tab:nice_patterns}.

 \begin{table}[]t]
 \small
     \centering \scriptsize
     \begin{tabular}{|c|c|c|c|c|c|} \hline
          $\delta$ & n. of genes & n. of cond. & area & MSR & $d$ \\ \hline
          $0.1$ & 6 & 5 & 30 & 0.00013 & 5.45 \\ 
          $0.2$ & 7 & 7 & 49 & 0.00130 & 7.00\\ 
          $0.3$ & 11 & 8 & 88 & 0.00487 & 9.26 \\ 
          $0.4$ & 12 & 9 & 108 & 0.00653 & 10.29\\ \hline 
     \end{tabular}
     \caption{Description of best $\delta$--shifting patterns (in terms of harmonic diameter) for several levels of $\delta$.}
     \label{tab:best_patterns}
 \end{table}

  \begin{table}[t]
  \small
     \centering \scriptsize
     \begin{tabular}{|c|c|c|c|c|c|} \hline
          $\delta$ & n. of genes & n. of cond. & area & MSR & $d$ \\ \hline
          $0.1$ & 5 & 5 & 25 & 0.00055 & 5.00 \\ 
          $0.2$ & 4 & 4 & 16 & 0.00151 & 4.00\\ 
          $0.3$ & 4 & 7 & 28 & 0.00137 & 5.09 \\ 
          $0.4$ & 4 & 9 & 36 & 0.00695 & 5.54\\ \hline 
     \end{tabular}
     \caption{Description of four $\delta$--shifting patterns (in terms of quality) for several levels of $\delta$.}
     \label{tab:nice_patterns}
 \end{table}
 
Biclustering has two remarkable properties that distinguishes it from clustering: a) rows or columns can belong to several biclusters, allowing overlapping; b) all the solutions together (final biclusters) do not need to completely cover the original matrix (dataset). The first property is visible in the behaviors present in Fig. \ref{fig:best_patterns}, in particular for $\delta=0.3$ and $\delta=0.4$. Biclusters only focus on patterns (locally) and not on partitions (globally), as clusters. This is also noticed in the solutions: the eight biclusters displayed in both figures use 61 rows (genes); however, there are only 27 unique rows in total. Irrelevant rows or columns would never appear in the solutions, satisfying the second property.

Finally, although it is out of the scope of this work to biologically analyse the quality of the patterns, it is a general attribute --- extendable to other domains --- that the behavior of genes (lines in the figures) present fluctuations, which might reveal up or down regulation. This attribute can be measured by the range coverage in terms of percentage, as this aspect is not usually illustrated in figures. For example, the bicluster displayed in Fig. \ref{fig:nice_patterns} for $\delta=0.3$ (bottom left) contains 7 genes (lines), representing the behavior of those genes for 4 experimental conditions. Each gene has a original range of values, and it is desirable that the pattern contains a great part of that range, as it would include fluctuations, instead of a flat behavior. The range coverage of that bicluster is, for each gene, 24.2\%, 23.5\%, 40\%, 31.7\%, 17.1\%, 30.7\% and 10.1\%, respectively, what indicates that only using 4 out of 9 experimental conditions the pattern is substantially varying across the original range.


 
 

    \begin{figure*}[!ht]
     \centering
     \includegraphics[width=\textwidth, trim = 110 30 90 10, clip]{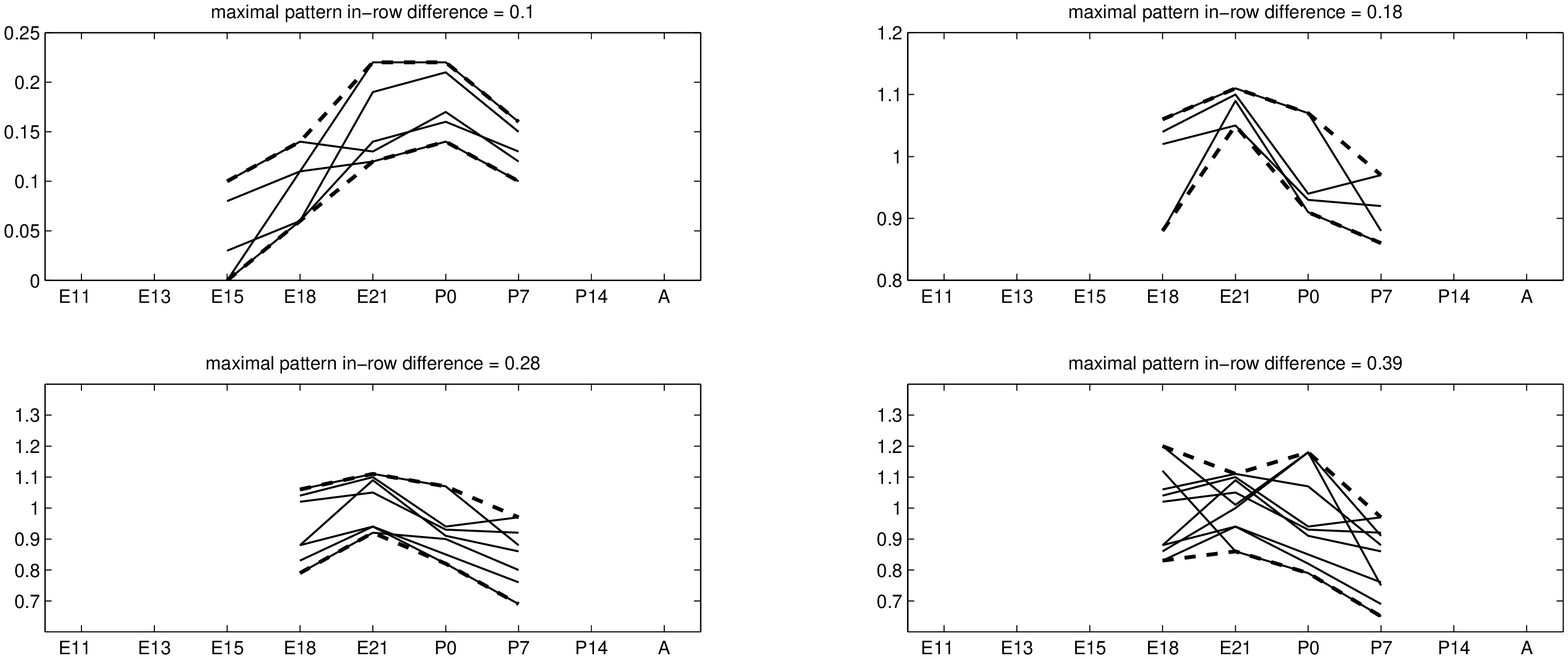}
     \caption{Additional $\delta$--shifting patterns found for several levels of $\delta$: $\delta = 0.1$ (upper left), $\delta = 0.2$ (upper right), $\delta=0.3$ (bottom left) and $\delta = 0.4$ (bottom right).}
     \label{fig:nice_patterns}
 \end{figure*}



\section{Conclusions}
In this work, the problem of biclustering on real--valued data from the Boolean reasoning perspective is addressed. Mathematical foundations are provided to support the use of Boolean concepts, in particular, the maximal--inclusion patterns, in the search for biclusters that include $\delta$--shifting patterns.

Whereas most algorithms use heuristic strategies to address the complexity of the biclustering problem and find biclusters that represent good (though not optimal) solutions, the presented approach always finds the optimal solutions for the $\delta$ threshold established for the $\delta$--shifting pattern.

Experiments on central nervous system development data suggest that the approach has excellent performance (measured by the mean squared residue) at finding large fluctuations of rows across columns (or vice versa) maintaining small fluctuation of columns (or rows). For example, in the case of gene expression analysis, large fluctuation of the values of each gene across the experimental conditions maintaining a small variation of the values of the subset of genes at each experimental condition.

Finally, future research direction will tackle the search of scaling patterns, for which it has been already proven that the mean squared residue is not an appropriate measure to score the quality of patterns, especially when combined with shifting patterns.

\section*{Acknowledgements}
The work was partially carried out within the statutory research grant of the Department of Computer Networks and Systems (RAu9 --- first author), and by the Spanish Ministry of Science, Innovation and Universities under project TIN2017--88209--C2--1--R and by the regional government of Junta de Andaluc\'ia (second author).

\bibliographystyle{abbrv}
\bibliography{mybibliography}

\section*{Appendix}
\subsection*{Proofs of Theorems for Constant Pattern Induction}\label{thms:exact}

\newtheorem{proof}{Proof of Theorem}
\begin{proof}
	\textbf{(\mytheoremonename)}  $\cali'\calj'$ is an implicant of $\ffunction(\calm)$ iff $\cali\calj$ is a constant pattern in $\calm$.
\end{proof}

\vspace*{.4cm}
\noindent $\Rightarrow$ Let $\cali'\calj'$ be an implicant of $\ffunction(\calm)$ and $\cali\calj$ be not a constant pattern in $\calm$. That means that there exists at least one pair of different columns $c,d \in \calj$ and a row $r \in \cali$ such that:
$$
v_{r,c} \neq  v_{r,d}
$$
This in turns means, that the clause $(r \lor c \lor d)$ is not satisfied by $\cali'\calj'$ what introduces the contradiction with the assumption.

\vspace*{.4cm}
\noindent $\Leftarrow$ Let $\cali\calj$ be a constant pattern in $\calm$ and $\cali'\calj'$ be not an implicant in $\ffunction(\calm)$. That means that there exists $r \in \cali$ and $c, d \in \calj$ such that 
$$
v_{r,c} \neq  v_{r,d} 
$$
what makes the contradiction.

\begin{proof}
	\textbf{(\mytheoremtwoname)} 	$\cali'\calj'$ is a prime implicant of $\ffunction(\calm)$ iff $\cali\calj$ is an inclusion--maximal constant pattern in $\calm$.
\end{proof}

\vspace*{.4cm}
\noindent $\Rightarrow$ Let $\cali'\calj'$ be a prime implicant of $\ffunction(\calm)$ and $\cali\calj$ be not an inclusion--maximal constant pattern in $\calm$. That means that there exists at least one column $c \in \calc\setminus \calj$ or row $r \in \calr\setminus\cali$ such that $(\cali\cup \{r\})\calj$ or $\cali(\calj\cup\{c\})$ is also a constant pattern. However, on the basis of the Theorem \ref{th:exact:correctness} $(\cali'\setminus\{r\})\calj$ or $\cali'(\calj'\setminus \{c\})$ is also an implicant of $\ffunction(\calm)$ so $\cali'\calj'$ can`t be the prime implicant and it is in contradiction to the assumptions.

\ \\
\noindent $\Leftarrow$ Let $\cali\calj$ be an inclusion--maximal constant pattern in $\calm$ and $\cali'\calj'$ be not a prime implicant in $\ffunction(\calm)$. That would mean that there exists $r \in \cali'$ or $c \in \calj'$ such that  $(\cali'\setminus\{r\})\calj$ or $\cali'(\calj'\setminus \{c\})$ is also an implicant of $\ffunction(\calm)$. This in turn means (Theorem \ref{th:exact:correctness}) that $(\cali\cup \{r\})\calj$ or $\cali(\calj \cup \{c\})$ will be an inclusion--maximal constant pattern in $\calm$ what makes a contradiction.

\subsection*{Proofs of Theorems for $\delta$--shifting Pattern Induction}\label{thms:shift_patt}

\begin{proof}
	\textbf{(\mytheoremthreename)}
	$\cali'\calj'$ is an implicant of $\ffunction_\delta(\calm)$ iff $\cali\calj$ is a $\delta$--shifting pattern in $\calm$.
\end{proof}

\noindent $\Rightarrow$ Let $\cali'\calj'$ be an implicant of $\ffunction_\delta(\calm)$ and $\cali\calj$ be not a $\delta$--shifting pattern in $\calm$. That means that there exists at least one pair of different columns $c, d$ and a row $r$ such that:
$$
|v_{r,c} - v_{r,d}| > \delta
$$
This in turns means, that the clause $(r \lor c \lor d)$ is not satisfied by $\cali'\calj'$ what introduces the contradiction with the assumption.

\ \\
\noindent $\Leftarrow$ Let $\cali\calj$ be a $\delta$--shifting pattern in $\calm$ and $\cali'\calj'$ be not an implicant in $\ffunction_\delta(\calm)$. That means that there exists $r \in \cali$ and $c, d \in \calj$ such that 
$$
|v_{r,c} - v_{r,d}| > \delta
$$
what makes the contradiction.

Having proved the Theorem \ref{th1} it is possible to prove Theorem \ref{th2}.

\begin{proof}
	\textbf{(\mytheoremfourname)}
	$\cali'\calj'$ is a prime implicant of $\ffunction_\delta(\calm)$ iff $\cali\calj$ is an inclusion--maximal  $\delta$--shifting pattern in $\calm$.
\end{proof}

\vspace*{.4cm}
\noindent $\Rightarrow$ Let $\cali'\calj'$ be a prime implicant of $\ffunction_\delta(\calm)$ and $\cali\calj$ be not an inclusion--maximal $\delta$--shifting pattern in $\calm$. That means that there exists at least one column $c \in \calc \setminus \calj$ or row $r \in \calr\setminus \cali$ such that $(\cali\cup \{r\})\calj$ or $\cali(\calj\cup\{c\})$ is also a $\delta$--shifting pattern. However, on the basis of the Theorem \ref{th1} $(\cali'\setminus\{r\})\calj'$ or $\cali'(\calj'\setminus \{c\})$ is also an implicant of $\ffunction_\delta(\calm)$ so $\cali'\calj'$ can`t be the prime implicant and it is in contradiction to the assumptions.

\ \\
\noindent $\Leftarrow$ Let $\cali\calj$ be an inclusion--maximal $\delta$--shifting pattern in $\calm$ and $\cali'\calj'$ be not a prime implicant in $\ffunction_\delta(\calm)$. That would mean that there exists $r \in \cali'$ or $c \in \calc\calj'$ such that  $(\cali'\setminus\{r\})\calj$ or $\cali'(\calj'\setminus \{c\})$ is also an implicant of $\ffunction_\delta(\calm)$. This in turn means (Theorem \ref{th2}) that $(\cali\cup \{c\})\calj$ or $\cali(\calj \cup \{c\})$ will be a $\delta$--shifting pattern bicluster in $\calm$ what makes a contradiction.

\subsection*{Proofs of Theorems for Exhaustive Search}\label{thms:exh}

\begin{proof}
	\textbf{(\mytheoremfivename)}
	$\cali'\calj'\cala'_p$ is an implicant of $\ffunction_\alpha(\calm)$ iff $\cali\calj$ is a $\alpha_p$--shifting pattern in $\calm$.
\end{proof}

\vspace*{.4cm}
\noindent $\Rightarrow$ Let $\cali'\calj'\cala'_p$ be the implicant of the function $\ffunction_\alpha(\calm)$ and $\cali\calj$ be not the $\alpha_p$--shifting pattern in the matrix $\calm$. That means that there exists at least one pair of cells in some row in the pattern $\cali\calj$ that violates the criterion of maximal absolute in-row difference  equal to $\alpha_p$ that is not covered by the $\cali'\calj'\cala'_p$, which is contradiction.

\vspace*{.4cm}
\noindent $\Leftarrow$ Let $\cali\calj$ be the $\alpha_p$--shifting pattern in the matrix $\calm$ and $\cali'\calj'\cala'_p$ be not the implicant of the $\ffunction_\alpha(\calm)$ function. That would mean that there exists at least one clause in $\ffunction_\alpha(\calm)$ that $\cali'\calj'$ does not cover. This, in turn, means that this clause codes the pair of cells in one row of $\cali\calj$ that violates the $\alpha_p$ condition, which is contradiction with the assumption that $\cali\calj$ is the $\alpha_p$--shifting pattern.

\begin{proof}
	\textbf{(\mytheoremsixnamep)}
	$\cali'\calj'\cala'_p$ is a prime implicant of $\ffunction_\alpha(\calm)$ iff $\cali\calj$ is an inclusion--maximal  $\alpha_p$--shifting pattern in $\calm$.
\end{proof}

\vspace*{.4cm}
\noindent $\Rightarrow$ Let $\cali'\calj'\cala'_p$ be the prime implicant of the $\ffunction_\alpha(\calm)$ Boolean function. From the previously proved theorem, $\cali\calj$ is the $\alpha_p$--shifting pattern  in $\calm$. Let  $\cali\calj$ be not inclusion--maximal pattern. That means that there exists at least one of the following:
\begin{itemize}
	\item $r \in \calr\setminus \cali$ such that $(\cali \cup \{r\})\calj$ is also an  $\alpha_p$--shifting pattern,
	\item $c \in \calc\setminus\calj$ such that $\cali(\calj \cup \{c\})$ is also an $\alpha_p$--shifting pattern.
\end{itemize} 
Additionally, from the previous theorem $\cali'(\calj\setminus\{c\})$ or $(\cali'\setminus \{r\})B$ are the implicants of $\ffunction_\delta(\calm)$, which is in contradiction with the assumption that $\cali'\calj'\cala'_p$ is the prime implicant.

\vspace*{.4cm}
\noindent $\Leftarrow$ Let $\cali\calj$ be the inclusion--maximal $\alpha_p$--shifting pattern in the matrix $\calm$. From the previous theorem  $\cali'\calj'\cala'_p$ is the implicant of the $\ffunction_\alpha(\calm)$ Boolean function. Let $\cali'\calj'\cala'_p$ be not the prime one. It means that there exists at least one of two kinds of literals $r\in \cali'$ or $c \in \calj'$ that can be removed from the implicant. The shortened implicant is equivalent to one of two extended $\alpha_p$--shifting patterns:
\begin{itemize}
	\item $(\cali\cup\{r\})\calj$ or 
	\item $\cali(\calj\cup \{r\})$
\end{itemize} 
which is contradiction with the statement that $\cali\calj$ is the inclusion--maximal $\alpha_p$--shifting pattern.

\subsection*{Proofs of Theorems for Pruned Search}\label{thms:exh_lim}

\begin{proof}
	\textbf{(\mytheoremsevenname)}
	$\cali'\calj'\cala'_p$ is an implicant of $\ffunction_{\delta,\alpha}(\calm)$ iff $\cali\calj$ is a $\alpha_p$--shifting pattern in $\calm$  (or the constant pattern if $\cala'_p = \emptyset$).
\end{proof}

\vspace*{.4cm}
\noindent $\Rightarrow$ Let $\cali'\calj'\cala'_p$ be the implicant of the function $\ffunction_{\delta,\alpha}(\calm)$. Let $\cali\calj$ be not at most $\alpha_p$--shifting pattern (the constant pattern can be considered as ``at most'' any $\alpha$--shifting pattern. That would mean that there exists at leas one pair of different cells in $\cali\calj$ pattern rows that violates the $\alpha_p$ condition. However, in such a case they should be covered by the $\cali'\calj'\cala'_p$ implicant and they are not what makes the contradiction.

\vspace*{.4cm}
\noindent $\Leftarrow$ Let $\cali\calj$ be the $\alpha_p$--shifting pattern (the constant in particular) and $\cali'\calj'\cala'_p$ be not the implicant of $\ffunction_{\delta,\alpha}(\calm)$. That would mean that there exists pair of different cells in some $\cali\calj$ pattern row which absolute difference exceeds $\alpha_p$ what comes from the fact that such pair should not be covered by the $\cali'\calj'\cala'_p$ expression. That is in contradiction that $\cali\calj$ is $\alpha_p$--shifting pattern.

Concluding, it was proved that implicants (of any form) of $\ffunction_{\delta,\alpha}(\calm)$ encode constant or $\alpha_p$--shifting patterns in $\calm$.

\begin{proof}
	\textbf{(\mytheoremeightname)}
	$\cali'\calj'\cala'_p$ is a prime implicant of $\ffunction_{\delta,\alpha}(\calm)$ iff $\cali\calj$ is an inclusion--maximal  $\alpha_p$--shifting pattern in $\calm$  (or the inclusion--maximal constant pattern if $\cala'_p = \emptyset$).
\end{proof}

\vspace*{.4cm}
\noindent $\Rightarrow$ Let $\cali'\calj'\cala'_p$ be the prime implicant of the function $\ffunction_{\delta, \alpha}(\calm)$ and $\cali\calj$ be not the inclusion--maximal $\alpha_p$--shifting pattern in $\calm$. It implies from the theorem above that $\cali\calj$ satisfies the $\alpha_p$ criterion. If the $\cali'\calj'\cala'_p$ corresponding pattern is not inclusion maximal that would mean that there exists at least one row or one column that can be added to the pattern which will still be the $\alpha_p$ constant. It also comes from the previous theorem, that there corresponds exactly one implicant of the $\ffunction_{\delta, \alpha}(\calm)$ function. However, there should exists the implicant for the newly created extended pattern what means that the original implicant $\cali'\calj'\cala'_p$ was not the prime, what makes the contradiction.

\vspace*{.4cm}
\noindent $\Leftarrow$ Let $\cali\calj$ be the inclusion--maximal $\alpha_p$--shifting pattern and $\cali'\calj'\cala'_p$ be not the prime implicant of $\ffunction_{\delta, \alpha}(\calm)$. That would mean that there exists at least one row or column corresponding Boolean variable that may simplify the implicant. It implies from the theorem above that such simplified implicant encodes one constant (in terms of $\alpha_p$) pattern. It is also implied that the simplified implicant will encode the pattern wider (by row or column) than $\cali\calj$ what makes the contradiction with the assumption, that $\cali\calj$ is the inclusion--maximal.

\end{document}